\documentclass[conference]{IEEEtran}
\IEEEoverridecommandlockouts

\usepackage{cite}
\usepackage{amsmath,amssymb,amsfonts}
\usepackage{mathtools}
\usepackage[ruled,vlined,linesnumbered]{algorithm2e}
\usepackage{graphicx}
\usepackage{textcomp}
\usepackage[dvipsnames]{xcolor}
\usepackage{subcaption}
\usepackage[normalem]{ulem}
\usepackage{booktabs}
\usepackage{enumitem}
\usepackage{multirow}
\usepackage{todonotes}
\usepackage[shortcuts,acronym]{glossaries}

\SetArgSty{textnormal}

\newacronym{iid}{i.i.d.}{independent and identically distributed}
\newacronym{cicme}{CICME}{common and individual causal mechanism estimation}
\newacronym{ctl}{CTL}{Causal Transfer Learning}
\newacronym{ood}{OOD}{out-of-distribution}
\newacronym{fcm}{FCM}{functional causal model}
\newacronym{dag}{DAG}{directed acyclic graph}
\newacronym{anm}{ANM}{additive noise model}
\newacronym{dsf}{DSF}{differentiable score function}
\newacronym{shd}{SHD}{structural Hamming distance}
\newacronym{lshd}{LSHD}{local structural Hamming distance}
\newacronym{mse}{MSE}{mean squared error}
\newacronym{mlp}{MLP}{multilayer perceptron}

\DeclareRobustCommand{\IEEEauthorrefmark}[1]{\smash{\textsuperscript{\footnotesize #1}}}

\mathchardef\mhyphen="2D
\def\D{\mathcal D}
\def\R{\mathbb R}
\def\X{\mathbf X}
\def\x{\mathbf x}
\def\L{\mathcal L}
\def\s{\mathbf s}

\makeatletter
\newcommand{\removelatexerror}{\let\@latex@error\@gobble}
\makeatother

\begin{document}

\title{Causal Mechanism Estimation in Multi-Sensor Systems Across Multiple Domains}

\author{\IEEEauthorblockN{Jingyi Yu\IEEEauthorrefmark{1,2},
Tim Pychynski\IEEEauthorrefmark{1}, and 
Marco F. Huber\IEEEauthorrefmark{2,3}}
\IEEEauthorblockA{\IEEEauthorrefmark{1}Bosch Center for Artificial Intelligence, Renningen, Germany}
\IEEEauthorblockA{\IEEEauthorrefmark{2}Institute of Industrial Manufacturing and Management IFF, University of Stuttgart, Stuttgart, Germany}
\IEEEauthorblockA{\IEEEauthorrefmark{3}Fraunhofer Institute for Manufacturing Engineering and Automation IPA, Stuttgart, Germany \\
Email: \{jingyi.yu, tim.pychynski\}@de.bosch.com, marco.huber@ieee.org}}

\maketitle

\begin{abstract}
To gain deeper insights into a complex sensor system through the lens of causality, we present \ac{cicme}, a novel three-step approach to inferring causal mechanisms from heterogeneous data collected across multiple domains. By leveraging the principle of \ac{ctl}, \ac{cicme} is able to reliably detect domain-invariant causal mechanisms when provided with sufficient samples. The identified common causal mechanisms are further used to guide the estimation of the remaining causal mechanisms in each domain individually. The performance of \ac{cicme} is evaluated on linear Gaussian models under scenarios inspired from a manufacturing process. Building upon existing continuous optimization-based causal discovery methods, we show that \ac{cicme} leverages the benefits of applying causal discovery on the pooled data and repeatedly on data from individual domains, and it even outperforms both baseline methods under certain scenarios.

\end{abstract}


\section{Introduction}
\label{sec:intro}
In manufacturing systems, sensors are widely deployed to monitor machines, components, and environmental conditions. The data collected from the sensors are fused to provide accurate, robust, and consistent inferences about the state of the system. Machine learning-based fusion methods have gained widespread application due to their strong predictive capabilities. The successes largely stem from their ability to recognize patterns on suitably collected \ac{iid} data~\cite{scholkopf2021toward}. As a consequence, they often struggle with \ac{ood} generalizability when the test data exhibit a different distribution than the training data, making their deployment 
challenging. Moreover, due to their black-box nature, machine learning-based methods often lack the capabilities to provide insights into the modeled system and explain their predictive behavior. 

Unlike correlations, which are typically leveraged by machine learning methods, causality allows for predicting the effects of actions that perturb the observed system\cite{mooij2016distinguishing}. Uncovering the causal relationships among sensor measurements provides valuable insights into the interactions among system variables, enabling a better understanding of the system behavior. By distinguishing true causal effects from spurious correlations, fusion strategies based on causal knowledge can improve the robustness of prediction~\cite{yu2024causal, scholkopf2012causal, magliacane2018domain, arjovsky_invariant_2019} and accuracy of root cause analysis~\cite{yu2024causal}. 

One common approach to inferring causal relationships among variables is through the application of causal discovery methods, which typically estimate these relationships from observational data and represent them in the form of a \ac{dag}. Recent advances in causal discovery have introduced a continuous representation of \acp{dag}~\cite{zheng_dags_2018, zheng_learning_2020}, making it possible to search the space of possible \acp{dag} in a fully continuous regime using \acp{dsf} and gradient descent techniques. As data-driven approaches, most causal discovery methods assume \ac{iid} data,  meaning that each sample--comprising the values of a set of variables---is independently drawn from the same distribution. However, this assumption often does not hold in real-world applications, especially in complex manufacturing systems, where the collected sensor measurements often exhibit distribution shifts across factors such as the time of day, machine IDs, part types, and sensor types, among others.

We close the gap by extending the existing continuous optimization-based causal discovery methods to handle data from multiple domains where distribution shifts may arise. The domains can be defined flexibly based on specific aspects of the system one aims to understand. In other words, given any split of a dataset into multiple domains, we aim to identify where and how the behavior of the underlying sensor system changes across domains. Building upon the principle of \ac{ctl}~\cite{rojas2018invariant} by incorporating domain information into data, our approach is able to identify variables with invariant functional relationships to their direct causes, which we refer to as \emph{common causal mechanisms}, and use the identified common causal mechanisms shared by all domains to further guide the estimation of the remaining causal mechanisms in each individual domain.

The rest of the paper is organized as follows: Section~\ref{sec:preliminaries} gives a brief introduction to \acp{fcm}, causal mechanisms, and causal discovery. We formally describe our approach in Section~\ref{sec:methodology}. Section~\ref{sec:experiments} contains the experimental settings and results on simulated data. Finally, we conclude our work in Section~\ref{sec:conclusions}.

\section{Preliminaries}
\label{sec:preliminaries}
\subsection{Functional Causal Model and Causal Mechanisms}
\label{sec:identifiability}
An \ac{fcm} considers a set of observed variables $X_1, \dots, X_d$ associated with vertices of a \ac{dag}. Each observed variable is determined by an assignment
\begin{equation}
    X_j = f_j(\mathbf{PA}_j, N_j)~,\quad j=1,\dots,d~,
\label{eq:1}
\end{equation}
where $\mathbf{PA}_j$ stands for $X_j$'s parents in the graph and $N_j$ for an unobserved error due to omitted factors~\cite{pearl2009causality}. Parents are connected to $X_j$ by directed edges in the graph, representing direct causal relationships, as parents directly affect the assignment of $X_j$ through \eqref{eq:1}. Therefore, the graph is also referred to as a ``causal graph". We think of the function $f_j$ and the noise distribution as the \emph{causal mechanism} transforming direct causes $\mathbf{PA}_j$ into effect $X_j$~\cite{peters_elements_2017}. The set of noises $N_1, \dots, N_d$ is assumed to be jointly independent to ensure that the joint distribution entailed by \eqref{eq:1} can be factorized into causal conditionals
\begin{equation}
\label{eq:2}
    P(X_1,\dots,X_d) = \prod_{j=1}^dP(X_j\mid \mathbf{PA}_j)~.
\end{equation}
To facilitate the task of inferring the associated causal graph that describes the data generating process from data (a.k.a. \emph{causal discovery}), it is necessary to restrict the complexity with which causal mechanisms depend on noises. One common restriction is to assume the noise is additive, narrowing \eqref{eq:1} down to 
\begin{equation}
\label{eq:3}
    X_j = f_j(\mathbf{PA}_j) + N_j~.
\end{equation}

While functions $f_j$ in \eqref{eq:3} can take any form, the causal discovery community has focused on special cases to obtain identifiability results. Assuming additive noise, it has been proven that the causal graph can be identified from data in the following cases: a) the functions $f_j$ are linear and the noises $N_j$ are non-Gaussian~\cite{shimizu_linear_2006}, b) the functions $f_j$ are nonlinear~\cite{hoyer_nonlinear_2008} , and c) the functions $f_j$ are linear and the noises $N_j$ are Gaussian with equal variances~\cite{peters_identifiability_2014}.

\subsection{Continuous Optimization-based Methods}
\label{subsec:CD}
At the top level, causal discovery methods fall into two categories: \emph{constraint-based } methods and \emph{score-based} methods. Constraint-based methods, such as the PC~\cite{spirtes_causation_2000} and FCI~\cite{spirtes_fci_1999} algorithms, view causal structure learning as a constraint satisfaction problem, where conditional independencies inferred from observational data are used to iteratively prune the space of possible graphs. In contrast, score-based methods, such as GES~\cite{chickering_optimal_2002}, formulate it as a combinatorial optimization problem. They optimize a given score function, which quantifies how well the graph fits the data, with the constraint that the graph is a \ac{dag}. 

By leveraging a novel algebraic characterization of \acp{dag}, Zheng et al.~\cite{zheng_dags_2018, zheng_learning_2020} recast the score-based combinatorial optimization problem as a continuous optimization problem
\begin{equation}
\label{eq:4}
\min_{W\in\R^{d\times d}} F(W)\quad \text{subject to } h(W) = 0~,
\end{equation}
where the graph with $d$ nodes is defined by the weighted adjacency matrix $W\in\R^{d\times d}$, and $h:\R^{d\times d}\to\R$ is a differentiable function that measures the acyclicity of the graph such that $h(W)=0$ if and only if $W$ leads to a \ac{dag}. Consider a \ac{dsf} $F(W)$, the optimization problem can be solved using gradient descent methods. Recent advances in this line of work have been focusing on the deployment of different \acp{dsf}, \ac{dag} constraints, and modeling techniques to approximate the functional relationships among the variables ~\cite{Lachapelle2020Gradient-Based, Zhu2020Causal, zheng_learning_2020, ng_role_2020, he_daring_2021, bello_dagma_2022, deng2025markov}. To this point, we want to briefly review NOTEARS-MLP \cite{zheng_learning_2020}.

Given a data matrix $\X=[\x_1,\dots, \x_d]\in\R^{n\times d}$ consisting of $n$ \ac{iid} observations from the joint distribution $P(X)$, NOTEARS-MLP uses a \ac{mlp} to model the functional relationship $f_j$ of each variable $X_j\in\{X_1,\dots,X_d\}$. Consider an \ac{mlp} with $h$ hidden layers and a single activation function $\sigma:\R\to\R$ to approximate $f_j$ such that
\begin{equation}
\label{eq:5}
    \mathrm{MLP}(\X; \theta_j) = \sigma(A_j^{(h)}\sigma(\cdots\sigma(A_j^{(1)}\X^\top)))~,
\end{equation}
where $\theta_j = (A_j^{(1)},\dots A_j^{(h)})$ denotes the parameters of the \ac{mlp} and $A_j^{(\ell)}\in\R^{m_\ell\times m_{\ell-1}}$ corresponds to the weights of the $\ell\mhyphen\mathrm{th}$ layer. It is easy to see that $\mathrm{MLP}(\X;\theta_j)$ is independent of $X_k$ if the $k$th-column of $A_j^{(1)}$ only consists of zeros. This results in $[W(f)]_{kj}=0$ if $\Vert k\mathrm{th}\mhyphen\mathrm{column}(A_j^{(1)})\Vert_2=0$.

Let $\theta=(\theta_1,\dots,\theta_d)$ denote the parameters for the set of \acp{mlp}. The weighted adjacency matrix $W$ in \eqref{eq:4} can be extracted from $\theta$ according to 
\begin{equation}
\label{eq:6}
    [W(\theta)]_{kj}=\Vert k\mathrm{th}\mhyphen\mathrm{column}(A_j^{(1)})\Vert_2~.
\end{equation}
The optimization problem in \eqref{eq:4} is thus defined as
\begin{equation}
\label{eq:7}
\begin{gathered}
    \min_\theta\quad\frac{1}{n}\sum_{j=1}^d\ell(\x_j, \mathrm{MLP}(\X;\theta_j)) + \lambda\Vert A_j^{(1)}\Vert_{1} \\
    \text{subject to }h(W(\theta)):=\mathrm{tr}(e^{W(\theta)\circ W(\theta)})-d=0~,
\end{gathered}  
\end{equation}
where $\ell$ is a loss function such as the least squares or the negative log-likelihood, $\ell_1$ regularization $\Vert A_j^{(1)}\Vert_{1}$ is used to enforce sparsity in the learned \ac{dag}, and the acyclicity constraint is enforced by $h(W(\theta))$ where $\circ$ denotes the Hadamard product and $e^A$ is the matrix exponential of $A$. Using the machinery of augmented Lagrangian method~\cite{Bertsekas01031997}, the solution of the equality-constrained problem in \eqref{eq:7} can be well approximated by solving an unconstrained $\ell_1$-penalized smooth minimization problem with the L-BFGS-B algorithm~\cite{byrd1995limited}.

\section{Methodology}
\label{sec:methodology}
Built upon existing continuous optimization-based causal discovery methods (see Section~\ref{subsec:CD}), \ac{cicme} is proposed to recover the causal relationships among variables from multi-domain data, where domains can be defined as any factor of interest that may cause the change of causal mechanisms for certain variables. In the first step, we apply causal discovery methods on samples pooled from all domains. In the second step, it identifies stable variables whose causal mechanisms remain invariant across domains. In the third step, we apply causal discovery methods on samples from individual domains, guided by the common causal mechanisms recovered in the previous step. We describe our methodology as a specific instantiation of NOTEARS-MLP, but it is applicable to other approaches, such as GOLEM~\cite{ng_role_2020}, NOTEARS-Sobolev~\cite{zheng_learning_2020}, and DAGMA~\cite{bello_dagma_2022}.

\subsection{Step~$1$: Causal Discovery on Pooled Data}
\label{subsec:step1}
Given datasets $\D_k=\{\s_i^{(k)}\}_{i=1}^{n_k}$ from multiple domains $k\in\{1,\dots,K\}$, we create $(\X, D)=\bigcup_{k=1}^K\{(\s_i^{(k)}, k) \mid i=1,2,\dots,n_k\}$ by pooling samples from all datasets and including a discrete variable $k$, which indicates the domain index. In the first step, we apply NOTEARS-MLP on the pooled samples $\X$ to identify the causal structure on pooled data. We refer readers to Algorithm~\ref{alg:step1}.
\begin{figure}[!t]
 \removelatexerror
\begin{algorithm}[H]
    \caption{Pseudo-code for step~$1$}
    \label{alg:step1}
    \SetAlgoLined
    \DontPrintSemicolon
    \SetKwInput{Init}{Init}
    \SetKwInput{Input}{Input}
    \SetKwInput{Output}{Output}
    \Init{Parameters $\theta=(\theta_1, \dots,\theta_d)$ for NOTEARS-MLP}
    \Input{Datasets $\D_k=\{\s_i^{(k)}\}_{i=1}^{n_k}$ for $k\in\{1,\dots,K\}$}
    \Output{Parameters $\theta=(\theta_1, \dots,\theta_d)$}
    Pool datasets to form $(\X, D)=\bigcup_{k=1}^K\{(\s_i^{(k)}, k) \mid i=1,2,\dots,n_k\}$ \;
    \While{the loss is not converged}{
        Calculate $\L_{\mathrm{NOTEARS}}(\theta\mid \X)$ in \eqref{eq:7}\;
        Update $\theta$ to minimize the loss \;
        }
    \Return $\theta$
\end{algorithm}
\end{figure}

\subsection{Step~$2$: Detection of Stable Variables}
\label{subsec:step2}
To find stable variables whose local causal mechanism $f_j$ remains invariant across all domains, we adopt the idea from \ac{ctl}~\cite{rojas2018invariant}, which aims to find a set of predictors leading to invariant conditionals by checking whether the residuals of prediction models are independent of the domain indices using the nonparametric HSIC test~\cite{gretton2007kernel}. In particular, the test sample $Z=(R_{j,i},D_i)_{i=1}^n$ is drawn from a joint distribution over residuals of $\mathrm{MLP}(\X;\theta_j)$ and domain indices, where $n=\sum_{k=1}^Kn_k$ corresponds to the total number of samples from all domains. Let $\mathrm{HSIC}_b(Z)$ be the test statistic. Since its value under the null hypothesis can be approached by a Gamma distribution, the associated p-value $p^*$ can be approximated as 
\begin{equation}
\label{eq:8}
    p^*\approx1-F_\mathrm{Ga}(\mathrm{HSIC}_b(Z))~,
\end{equation}
where $F_\mathrm{Ga}$ is the cumulative distribution function of the gamma distribution.
The null hypothesis $H_0$ of the independence between residuals and domain indices is rejected if $p^*$ is below the significance level $\alpha$, otherwise we fail to reject it, in which case we claim that the estimated causal mechanism $f_j$ is domain-invariant. As a result, we find the stable variables with invariant causal mechanisms by collecting all \acp{mlp} whose residuals are independent of domain indices. We provide the pseudo-code of this step in Algorithm~\ref{alg:step2}. 

\begin{figure}[!t]
 \removelatexerror
\begin{algorithm}[H]
    \caption{Pseudo-code for step~$2$}
    \label{alg:step2}
    \SetAlgoLined
    \DontPrintSemicolon
    \SetKwInput{Init}{Init}
    \SetKwInput{Input}{Input}
    \SetKwInput{Output}{Output}
    \Input{Pooled data $(\X, D)$, significance level $\alpha$ for independence test}
    \Output{Stable variable(s) $X_\mathrm{stable}$}
    
    Run Algorithm~\ref{alg:step1} to get $\theta=(\theta_1, \dots,\theta_d)$\;
    Set $X_\mathrm{stable}=\{\}$ \;
    \For{$\theta_j\in\{{\theta_1,\dots,\theta_d\}}$}{
        Compute the residuals $R_j$ using $\mathrm{MLP}(\X;\theta_j)$ \;
        Compute $\mathrm{HSIC}_b(R_j,D)$ and associated p-value $p^*$ \;
        \If{$p^*>\alpha$}{
            Add node $X_j$ to $X_\mathrm{stable}$ \;
        }
    }

    \Return $X_\mathrm{stable}$
\end{algorithm}
\end{figure}

\subsection{Step~$3$: Causal Discovery on Individual Datasets}
\label{subsec:step3}
After detecting stable variables with invariant causal mechanisms, we recover the remaining domain-specific causal mechanisms by applying NOTEARS-MLP with a random initialization in each individual domain. To consistently recover the causal structure of the stable variables in each domain, we propose two alternative approaches to limiting the degrees of freedom in updating the weights of stable \acp{mlp}:
\begin{itemize}
    \item We freeze the weights of stable \acp{mlp} by restoring their values before training and set their gradients to zero during training, as illustrated in Algorithm~\ref{alg:step3_f}. We refer to this variant as \emph{\mbox{\ac{cicme}-f}}.
    \item We tailor the original loss function by adding an additional penalization term that promotes consistent recovery of the common causal structure obtained in the previous step, as illustrated in Algorithm~\ref{alg:step3_l}. We refer to this variant as \emph{\mbox{\ac{cicme}-l}}.
\end{itemize}

\begin{figure}[!t]
 \removelatexerror
\begin{algorithm}[H]
    \caption{Pseudo-code for step~$3$ (\mbox{\ac{cicme}-f})}
    \label{alg:step3_f}
    \SetAlgoLined
    \DontPrintSemicolon
    \SetKwInput{Init}{Init}
    \SetKwInput{Input}{Input}
    \SetKwInput{Output}{Output}
    \Init{Parameters $\theta^{(k)}=(\theta_1^{(k)}, \dots,\theta_d^{(k)})$ for $k\in\{1,\dots,K\}$}
    \Input{Datasets $\D_k=\{\s_i^{(k)}\}_{i=1}^{n_k}$ for $k\in\{1,\dots,K\}$}
    \Output{Individual causal mechanisms $\theta^{(k)}$ for $k\in\{1,\dots,K\}$}

    Run Algorithm~\ref{alg:step1} and \ref{alg:step2} to get $\theta$ and $X_\mathrm{stable}$\;
    \For{$k\in\{1,\dots,K\}$}{
        \For{$j\in\{1,\dots,d\}$}{
            \If{$X_j\in X_\mathrm{stable}$}{
            Load the parameters $\theta^{(k)}_j=\theta_j$\;
            }
        }

    }
    \While{the loss is not converged for all $k$}{
        \For{$k\in\{1,\dots,K\}$}{
             Calculate $\L_{\mathrm{NOTEARS}}(\theta^{(k)}\mid \s^{(k)}\sim\D_k)$ in \eqref{eq:7} \;
             \For{$j\in\{1,\dots,d\}$}{
                \If{$X_j\in X_\mathrm{stable}$}{
                    Set the gradients of $\theta^{(k)}_j$ to zero\;
                }
            }
             Update $\theta^{(k)}$ to minimize the loss\;
        }
    }
\end{algorithm}
\end{figure}

\begin{figure}[!t]
 \removelatexerror
\begin{algorithm}[H]
    \caption{Pseudo-code for step~$3$ (\mbox{\ac{cicme}-l})}
    \label{alg:step3_l}
    \SetAlgoLined
    \DontPrintSemicolon
    \SetKwInput{Init}{Init}
    \SetKwInput{Input}{Input}
    \SetKwInput{Output}{Output}
    \Init{Parameters $\theta^{(k)}=(\theta_1^{(k)}, \dots,\theta_d^{(k)})$ for $k\in\{1,\dots,K\}$}
    \Input{Datasets $\D_k=\{\s_i^{(k)}\}_{i=1}^{n_k}$ for $k\in\{1,\dots,K\}$}
    \Output{Individual causal mechanisms $\theta^{(k)}$ for $k\in\{1,\dots,K\}$}
    Run Algorithm~\ref{alg:step1} and \ref{alg:step2} to get $\theta$ and $X_\mathrm{stable}$ \;
    Create $M\in\R^{d\times d}$ for common causal mechanisms \;
    \While{the loss is not converged for all $k$}{
        \For{$k\in\{1,\dots,K\}$}{
            Calculate $\L_{\mathrm{NOTEARS}}(\theta^{(k)}\mid \s^{(k)}\sim\D_k)$ in \eqref{eq:7}\;
            Calculate $\L_\mathrm{com}$ in \eqref{eq:9}\;
            $\L\gets\L_{\mathrm{NOTEARS}}+\gamma\L_\mathrm{com}$\;
            Update $\theta^{(k)}$ to minimize the loss \;
        }
    }
\end{algorithm}
\end{figure}

The additional loss term adopted by \mbox{\ac{cicme}-l} is essentially the \ac{mse} between the predicted edge weights of stable variables using the pooled samples and samples from individual domains
\begin{equation}
\label{eq:9}
    \L^{(k)}_\mathrm{com}=\frac{1}{\sum_{i,j}M_{ij}}\sum_{i,j}M_{ij}(W_{\mathrm{pool},ij}-W^{(k)}_{\mathrm{ind,}ij})^2~,
\end{equation}
where $M$ is a mask matrix such that $M_{:j}=1$ if $X_j$ is a stable variable. $W_\mathrm{pool}$ and $W^{(k)}_\mathrm{ind}$ are the adjacency matrices estimated from the pooled data and data from the $k\mhyphen th$ domain, respectively. The overall loss function employed by NOTEARS-MLP in the $k\mhyphen th$ domain is thus
\begin{equation}
\label{eq:10}
    \L=\L_{\mathrm{NOTEARS}}(\theta^{(k)}\mid \s^{(k)}\sim\D_k) + \gamma\\L^{(k)}_\mathrm{com}~,
\end{equation}
where $\gamma$ is a non-negative weighting factor that controls how strongly the model aligns with the common causal structure.

\section{Experiments}
\label{sec:experiments}
In this section, we generate synthetic data comprised of multiple domains based on linear Gaussian models to compare the performance of \ac{cicme} and baseline methods. To assess \ac{cicme}'s ability to detect common causal mechanisms, we consider $4$ scenarios inspired from a real manufacturing process where causal mechanisms may shift across domains. 
\subsection{Baselines}
Without loss of generality, we focus on NOTEARS-MLP\footnote{We use the implementation from https://github.com/xunzheng/notears.} as in Section~\ref{sec:methodology} to demonstrate the benefits and limitations of \ac{cicme}. We choose the following two strategies as baselines:
\begin{itemize}
    \item Applying NOTEARS-MLP on the pooled dataset, referred to as \emph{NOTEARS-pool}. It is equivalent to step~$1$ of \ac{cicme}.
    \item Applying NOTEARS-MLP in each domain individually, referred to as \emph{NOTEARS-ind}. It is equivalent to step~$3$ of \ac{cicme} with $X_\mathrm{stable}$ being an empty set.
\end{itemize}

\subsection{Metrics}
To evaluate the capability of \ac{cicme} to identify common causal mechanisms that remain invariant in all domains, we report the total number of cases in which each variable is identified as stable in each experiment. In addition, the local causal structure of the stable variables is evaluated using a \ac{lshd}, which is $0$ if the causal parents are correctly identified. 

To assess whether the correct causal structure is found by \ac{cicme} and the baseline methods, we compare the estimated causal structure with the ground truth graph using the \ac{shd}, which counts the total number of edge additions, deletions, or reversals that are required to convert one graph into another. A lower \ac{shd} indicates that the estimated causal structure is closer to the ground truth. For each domain $k\in\{1,\dots,K\}$, the weighted adjacency matrix $W(\theta^k)$ is inferred from $\theta^{(k)}$ using \eqref{eq:6}. To obtain the estimated causal structure, a post-processing threshold of $0.3$ is applied on $W(\theta^k)$ to remove all edges with small weights, following the methodology in the NOTEARS papers\cite{zheng_dags_2018, zheng_learning_2020}. The reported \ac{shd} is averaged over all domains.

Since the execution time reflects the intrinsic computational complexity and the relative complexity of all methods remains unchanged across experiments, we choose to explicitly report the execution time of all methods on \ref{exp:1}.

\subsection{Scenarios}
\label{sec:scenarios}
\begin{figure}
    \centering
    \includegraphics[width=0.5\linewidth]{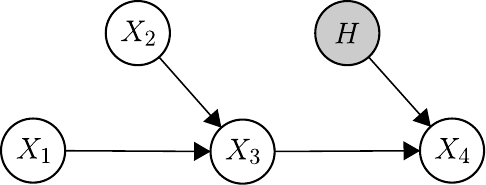}
    \caption{Causal graph of the variables in a leakage test process.}
    \label{fig:toyFCM}
\end{figure}

We consider the ground truth causal graph shown in Fig.~\ref{fig:toyFCM} for the generation of simulation data to evaluate our methods. It is based on a real manufacturing process---leakage test---that applies air pressure to a chamber and monitors the pressure drop to assess whether leakage occurs. In this process, $X_1$, $X_2$, $X_3$, and $X_4$ measure the chamber temperature, the air flow rate into the chamber, the chamber pressure before the test, and the pressure after a certain period, receptively. The pressure before the test $X_3$ is determined by both temperature $X_1$ and flow rate $X_2$, as higher temperature or increased air supply leads to higher pressure. After the test, the pressure $X_4$ is directly affected by $X_3$, with additional unmeasured factors such as the leakage area of the chamber, represented by $H$. Larger leakage areas cause the pressure to drop more quickly, and in extreme cases, the pressure may even reach the ambient pressure. 

In the context of conducting leakage tests across different product batches on the manufacturing line, several scenarios may arise, affecting the causal mechanisms among the measured variables:
\begin{enumerate}[label=(S{{\arabic*}})]
\item \emph{Variation in Leakage Area Across Batches}: The leakage area $H$ differs across product batches, altering the causal mechanism between the chamber pressure before the test $X_3$ and after the test $X_4$.
\item \emph{Chamber Pressure Reaching Ambient Pressure}: An anomaly in the preceding process, such as improper sealing of the chamber, may result in the leakage area $H$ becoming the primary factor in determining $X_4$, making the pressure before the test $X_3$ irrelevant. 
\item \emph{Flow Rate with Shifted Distributions}: The air flow rate $X_2$ typically depends on the states of the air supply system and thus fluctuates around a fixed value, which may be set differently based on different product types across batches. 
\item \emph{Flow Rate as Fixed Process Control Parameter}: The air flow rate $X_2$ in a well-controlled air supply system can be considered as a constant process control parameter. While it remains fixed within a given batch, it may be adjusted to different fixed values for various product types.
\end{enumerate}

\subsection{Experimental Setup}
Adhering to the additive \ac{fcm} defined in \eqref{eq:3}, we assume all other unmeasured causes of each measured variable $X_i$ are represented by an additive noise term $N_i$. Based on the causal structure defined in Fig.~\ref{fig:toyFCM}, we formalize the functions of measured variables as 
\begin{equation}
\label{eq:11}
\begin{aligned}
X_1 &:= N_1, \\ 
X_2 &:= N_2, \\
X_3 &:= X_1 + X_2 + N_3, \\
X_4 &:= H \times X_3 + N_4.
\end{aligned}
\end{equation} 
To ensure identifiability results as discussed in Section~\ref{sec:identifiability}, we assume the independent noise variables $N_1$, $N_2$, $N_3$, and $N_4$ follow standard Gaussian distributions, unless stated otherwise. Data from multiple domains are generated by sampling from domain-specific \acp{fcm}, each defined differently according to \eqref{eq:11}.

Inspired by the aforementioned scenarios in manufacturing where the causal mechanisms among measured variables may change across domains, a comprehensive set of experiments are conducted to compare \ac{cicme} with baseline methods using the following setups:
\begin{enumerate}[label=(E{{\arabic*}})]
    \item \emph{Varied Causal Mechanism Across Domains}: We generate three datasets $\D_1$, $\D_2$, and $\D_3$ with sample size $n\in\{10, 10^2, 10^3\}$, where $H$ takes on a fixed value sampled from $[-2, -0.5]\cup[0.5, 2]$, the same range of weight coefficients as used by NOTEARS~\cite{zheng_dags_2018}. The range $(-0.5, 0.5)$ is skipped to make sure the edge $X_3\to X_4$ is identifiable from data. \label{exp:1}
    \item \emph{Varied Causal Structure Across Domains}: We change the data generation process in \ref{exp:1} by assigning $0$ to $H$ in a randomly chosen domain, effectively cutting off the edge between $X_3$ and $X_4$. \label{exp:2}
    \item \emph{Shifted Distributions of Independent Variable}: We generate three datasets $\D_1$, $\D_2$, and $\D_3$ with sample size $n\in\{10, 10^2, 10^3\}$ and $H=1$. We shift the distribution of $X_2$ by sampling a random value from $[-2, -0.5]\cup[0.5, 2]$ as the mean for $N_2$ in each dataset.\label{exp:3}
    \item \emph{Shifted Values of Independent Variable}: We change the data generation process in \ref{exp:3} by setting $N_2$ to a fixed value sampled from $[-2, -0.5]\cup[0.5, 2]$ for each dataset. \label{exp:4}

\end{enumerate}
All of the above experiments are repeated $100$ times. Following the experimental setup in the paper~\cite{zheng_learning_2020}, we use \acp{mlp} with one hidden layer with 10 units and sigmoid activation function. Based on empirical results, we set $\gamma=10$ to scale the additional loss term in~\eqref{eq:10}. Consistent with previous work, we do not perform any hyperparameter optimization to avoid unintentionally biased results. We use the same $\lambda=0.01$ for all methods since our primary concern is whether CICME improves the baseline methods under the same condition.
\subsection{Results}
We first examine the capability of \ac{cicme} to identify common causal mechanisms across all domains based on the HSIC test results from step~$2$. Note that in \ref{exp:1} and \ref{exp:2}, $X_4$ exhibits changing causal mechanisms due to varying values of $H$ across domains, while in \ref{exp:3} and \ref{exp:4}, $X_2$ is the expected unstable variable due to different distributions of $N_2$ across domains. Table~\ref{tab:hsic} summarizes the total number of cases (out of $100$) in which each variable is identified as a stable variable. Additionally, the \ac{lshd} is averaged over those cases to indicate the accuracy of the estimated local structure. It can be seen that, when provided with $1000$ samples, \ac{cicme} is able to reliably identify variables with changing causal mechanisms, as evidenced by the rare cases of misclassifications highlighted in bold. However, they become more frequent with a reduced sample size, since fewer samples provide less information about the population, leading to less statistical power of the HSIC test. Similarly, the accuracy of the estimated causal parents of stable variables also deteriorates when provided with fewer samples, as illustrated by the higher \ac{lshd}. This may adversely impact the performance of \ac{cicme} by carrying over inaccurate common causal mechanisms to step~$3$.

In addition, we also observe that stable variables are occasionally identified as unstable. While this leads to higher degrees of freedom in updating the weights of the model, and hence, more training effort in step~$3$, it typically does not affect the overall performance. However, an exception is observed in \ref{exp:4}, where $X_3$ is identified as unstable significantly more often than stable even for $1000$ samples, despite the ground truth causal mechanism of $X_3$ being the same across all domains (i.e., $X_3=X_1+X_2+N_3$). We suspect that this is due to convergence issues in the constrained loss function. This can be hypothetically solved by finetuning the \ac{mlp} based on the least squares loss before applying the HSIC test in step~$2$. 

\begin{table*}[t]
  \caption{Identified stable variables based on the HSIC test results. The respective variable with changing causal mechanisms across domains is highlighted in bold for each experimental setup, whose stable count is expected to be $0$. The stable count for the other variables is expected to be $100$.}
  \label{tab:hsic}
  \centering
    \begin{tabular*}{\textwidth}{@{\extracolsep{\fill}}ll*{8}{c}}
        \toprule 
        \multirow{2}{*}{Experiment} & \multirow{2}{*}{Sample size} &
        \multicolumn{2}{c}{$X_1$} & \multicolumn{2}{c}{$X_2$} & \multicolumn{2}{c}{$X_3$} & \multicolumn{2}{c}{$X_4$}\\ 
        \cmidrule(lr){3-4} \cmidrule(lr){5-6} \cmidrule(lr){7-8} \cmidrule(lr){9-10}
        & & Stable count & \ac{lshd} & Stable count & \ac{lshd} & Stable count & \ac{lshd} & Stable count & \ac{lshd} \\
        \midrule
        \multirow{3}{*}{\ref{exp:1}} & $1000$ & $95$ & $0.00$ & $96$ & $0.00$ & $96$ & $0.00$ & $\mathbf{5}$ & $0.00$ \\
        & $100$ & $92$ & $0.01$ & $96$ & $0.01$ & $96$ & $0.00$ & $\mathbf{15}$ & $0.00$ \\
        & $10$ & $98$ & $0.20$ & $95$ & $0.28$ & $94$ & $0.04$ & $\mathbf{95}$ & $1.86$ \\
        \midrule
        \multirow{3}{*}{\ref{exp:2}} & $1000$ & $97$ & $0.00$ & $96$ & $0.00$ & $92$ & $0.00$ & $\mathbf{2}$ & $0.00$ \\
        & $100$ & $93$ & $0.00$ & $97$ & $0.00$ & $98$ & $0.00$ & $\mathbf{12}$ & $0.25$ \\
        & $10$ & $98$ & $0.28$ & $98$ & $0.19$ & $100$ & $0.10$ & $\mathbf{94}$ & $1.72$ \\
        \midrule
        \multirow{3}{*}{\ref{exp:3}} & $1000$ & $95$ & $0.00$ & $\mathbf{0}$ & $-$ & $93$ & $0.00$ & $92$ & $0.00$ \\
        & $100$ & $92$ & $0.02$ & $\mathbf{2}$ & $0.00$ & $99$ & $0.04$ & $96$ & $0.03$ \\
        & $10$ & $98$ & $0.07$ & $\mathbf{22}$ & $0.27$ & $94$ & $0.18$ & $97$ & $0.27$ \\
        \midrule
        \multirow{3}{*}{\ref{exp:4}} & $1000$ & $95$ & $0.00$ & $\mathbf{0}$ & $-$ & $37$ & $0.00$ & $89$ & $0.00$ \\
        & $100$ & $92$ & $0.00$ & $\mathbf{0}$ & $-$ & $92$ & $0.02$ & $96$ & $0.03$ \\
        & $10$ & $98$ & $0.17$ & $\mathbf{2}$ & $2.00$ & $90$ & $0.04$ & $99$ & $1.08$ \\
        \bottomrule
\end{tabular*}
\end{table*}

\subsubsection{Causal Structure}
\begin{figure}[bt!]
    \centering
    \begin{subfigure}{0.8\columnwidth}
      \includegraphics[width=\linewidth]{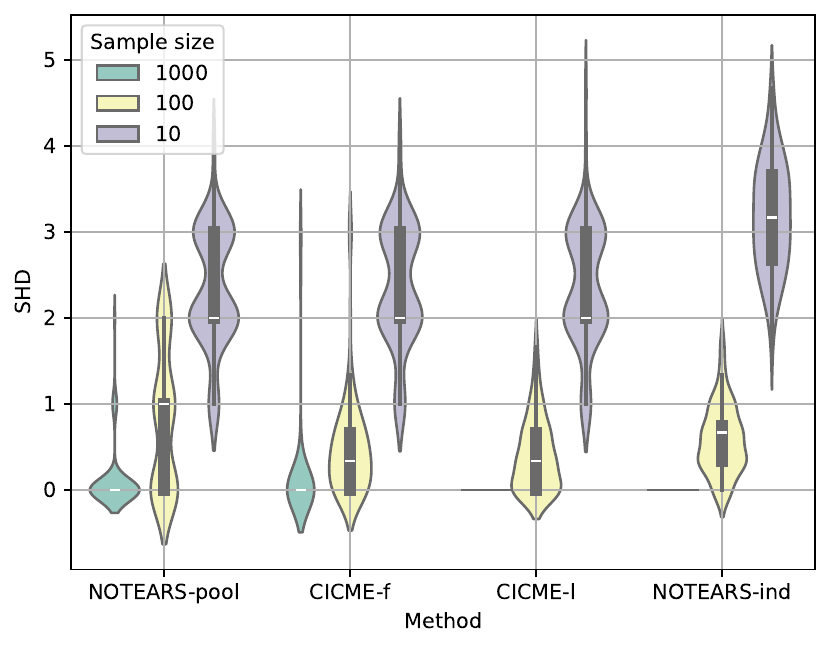}  
      \caption{The same causal structure---\ref{exp:1}}
      \label{fig:shd_exp1}
    \end{subfigure}
    \hfill
    \begin{subfigure}{0.8\columnwidth}
      \includegraphics[width=\linewidth]{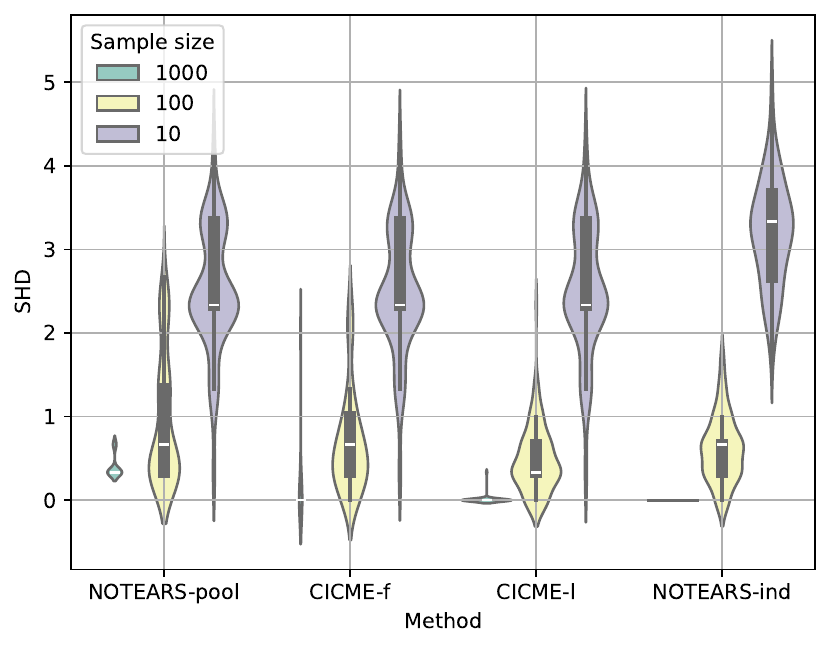}  
      \caption{Changing causal structures---\ref{exp:2}}
      \label{fig:shd_exp2}
    \end{subfigure}
    \caption{\ac{shd} averaged over all domains achieved by all methods when the causal mechanism of dependent variable $X_4$ changes across domains.}
    \label{fig:shd_12}
\end{figure}

We compare \ac{cicme} with the baseline methods in terms of \ac{shd} to evaluate whether the correct causal structure for each domain is found. Fig.~\ref{fig:shd_12} presents the averaged \ac{shd} of all methods when $X_4$ has changing causal mechanisms across domains. It is noticeable that regardless of whether the causal structure remains intact, NOTEARS-ind is likely to outperform NOTEARS-pool when provided with a sufficient number of samples (i.e., $1000$ and $100$), whereas the advantage of NOTEARS-pool becomes evident for a limited number of samples, as it leverages more data to infer the causal structure. Combining the strategy of both baseline methods, \mbox{\mbox{\ac{cicme}-l}} performs on par with the best of NOTEARS-pool and NOTEARS-ind for $1000$ and $10$ samples, and even surpasses both baseline methods for $100$ samples. With $1000$ and $100$ samples, \mbox{\ac{cicme}-l} outperforms \mbox{\mbox{\ac{cicme}-f}}, as it is likely to achieve a lower \ac{shd} in both cases. This suggests that it is desirable to leave more degrees of freedom by allowing the weights of stable \acp{mlp} to be updated when enough samples are available.

\begin{figure}[bt!]
    \centering
    \begin{subfigure}{0.8\columnwidth}
      \includegraphics[width=\linewidth]{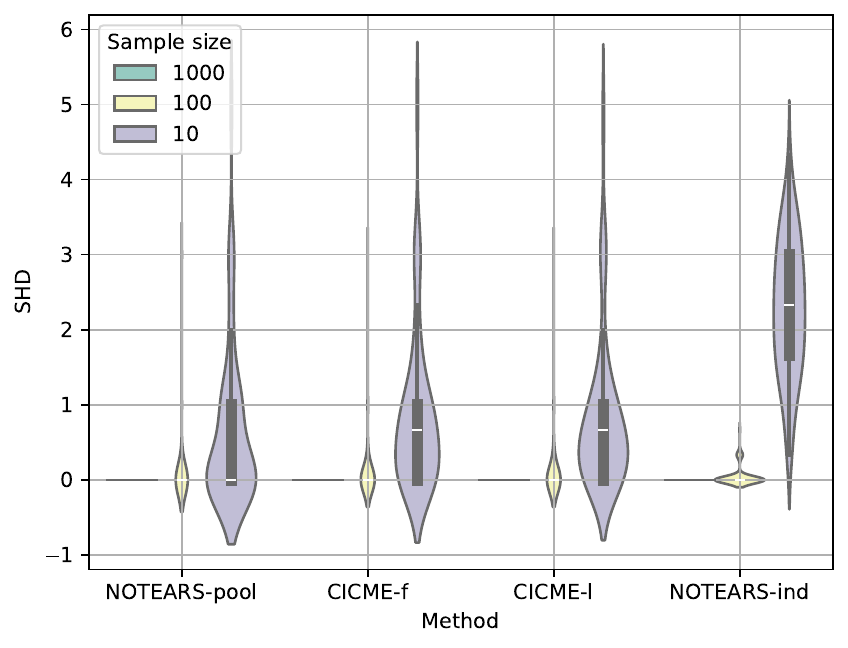}  
      \caption{Shifted distributions of $X_2$---\ref{exp:3}}
      \label{fig:shd_exp3}
    \end{subfigure}
    \hfill
    \begin{subfigure}{0.8\columnwidth}
      \includegraphics[width=\linewidth]{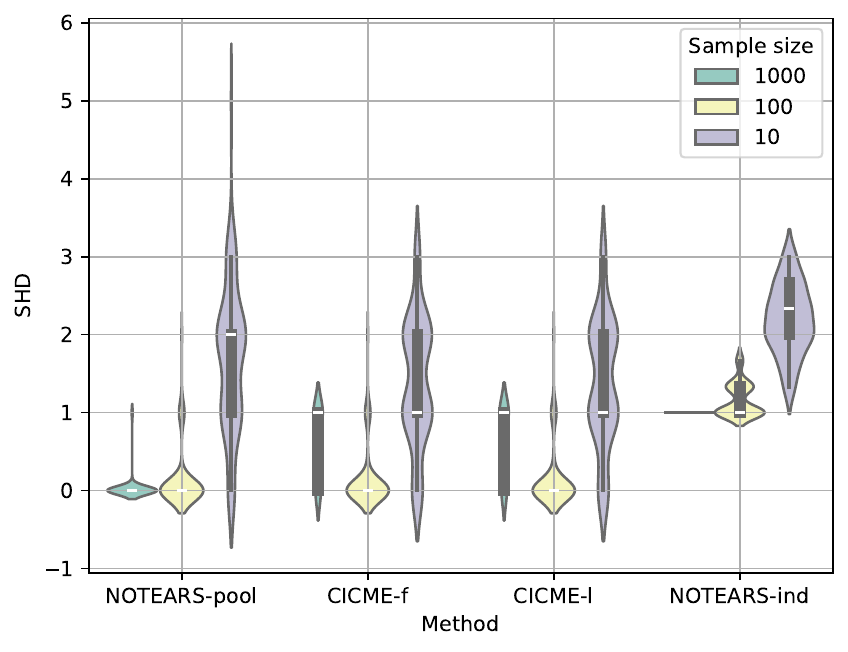}  
      \caption{Shifted values of $X_2$---\ref{exp:4}}
      \label{fig:shd_exp4}
    \end{subfigure}
    \caption{\ac{shd} averaged over all domains achieved by all methods when the causal mechanism of independent variable $X_2$ changes across domains.}
    \label{fig:shd_34}
\end{figure}

Moreover, Fig.~\ref{fig:shd_34} compares the capability of all methods to recover causal structures when the causal mechanism of $X_2$ changes across domains. When $X_2$ follows a distribution with shifted means across domains, the performance of \ac{cicme} is on par with either of the baseline methods, as shown in Fig.~\ref{fig:shd_exp3}. Similarly, NOTEARS-pool again outperforms NOTEARS-ind for $10$ samples, for the same reason mentioned earlier. However, this trend changes as soon as $X_2$ is fixed in each domain. Fig.~\ref{fig:shd_exp4} illustrates that NOTEARS-pool significantly outperforms NOTEARS-ind regardless of the sample size. This expected behavior can be attributed to the variance of $X_2$ in the observed data: The variance of $X_2$ equals to zero in each individual dataset and becomes non-zero in the pooled dataset, making it possible to infer the causal relationship between $X_2$ and other variables. Due to the pooling operation in step~$1$, \ac{cicme} is also able to outperform NOTEARS-ind regardless of the sample size, and it even achieves the best performance among all when the sample size is $10$. However, for a sample size of $1000$, \ac{cicme} fails to match the performance of the best of the baseline methods, in this scenario, NOTEARS-pool. We look for plausible explanations by investigating the results of the HSIC test in Table~\ref{tab:hsic}. We conclude that the inferior performance of \ac{cicme} is caused by the false identification of $X_3$ as unstable in most cases. If the causal mechanism of $X_3$ is not carried over to step~$3$, the causal edge $X_2\rightarrow X_3$ cannot be recovered because there is no variation in $X_2$ in each individual domain. This also partially explains the similar performance of both variants of \ac{cicme} in this scenario.

\subsubsection{Execution Time}
\begin{figure}[tb]
    \begin{subfigure}{\columnwidth}
      \centering
      \includegraphics[width=0.8\textwidth]{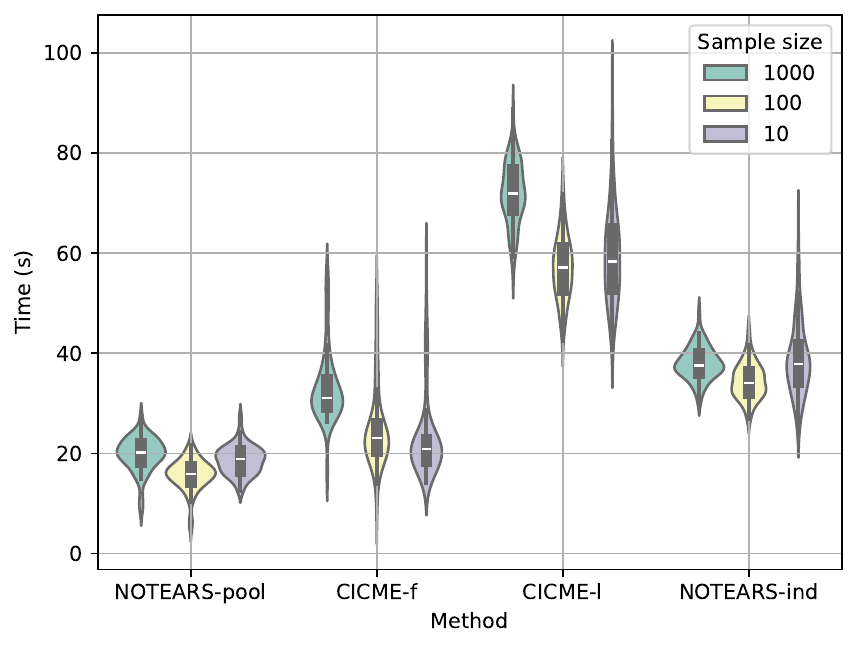}  
      \caption{Overall execution time}
      \label{fig:time_toyFCM}
    \end{subfigure}
    \hfill
    \begin{subfigure}{\columnwidth}
      \centering
      \includegraphics[width=0.8\textwidth]{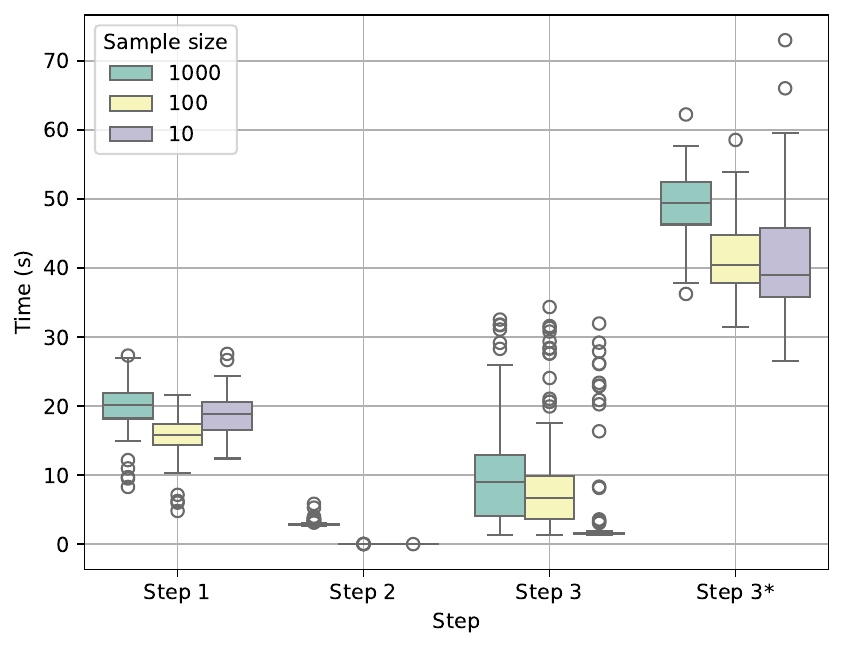} 
      \caption{Stepwise execution time of \ac{cicme}}
      \label{fig:time_CICME}
    \end{subfigure}
    \caption{Comparison of the execution time with different sample sizes. Step~$3$ (f) and Step~$3$ (l) in Fig~\ref{fig:time_CICME} denotes the execution time of step~$3$ of \mbox{\ac{cicme}-f} and \mbox{\ac{cicme}-l}, respectively. The box plot was chosen over the violin plot in Fig.~\ref{fig:time_CICME} for better color visualization.} 
    \label{fig:time}
\end{figure}

We compare the overall execution time of different methods in Fig.~\ref{fig:time_toyFCM} based on \ref{exp:1}. Surprisingly, the execution time is not always positively correlated with the sample size. For instance, NOTEARS-pool and NOTEARS-ind take longer to run on 10 samples than on 100 samples. This can be attributed to the fact that the model sometimes requires more epochs to reach convergence when trained on fewer samples. Among all methods, NOTEARS-pool is most time efficient since it only involves the optimization of NOTEARS-MLP once. Interestingly, by freezing the weights of stable \acp{mlp}, \mbox{\ac{cicme}-f} requires less execution time than NOTEARS-ind and, for small sample sizes, its execution time is only slightly higher than that of NOTEARS-pool, despite of the three-step approach. As expected, \mbox{\ac{cicme}-l} takes the longest time due to the computational overhead created by the additional loss term. A closer look at the execution time of each step of \ac{cicme} in Fig.~\ref{fig:time_CICME} reveals that the execution time of HSIC test (step~$2$) is almost negligible, whereas applying NOTEARS-MLP takes relatively longer time. Freezing weights of stable \acp{mlp} greatly facilitates the estimation of the remaining causal mechanisms, indicated by the difference between step~$3$ in Fig.~\ref{fig:time_CICME} and NOTEARS-ind in Fig.~\ref{fig:time_toyFCM}, further supporting the computational efficiency of \mbox{\ac{cicme}-f}.

\section{Conclusions}
\label{sec:conclusions}
In this paper, we have proposed a novel approach, \ac{cicme}, for inferring causal mechanisms from data collected across multiple domains. The definition of each domain can be determined flexibly, depending on the aspects of the system one aims to understand. In the context of manufacturing, meta information recorded in the systems, such as machine IDs, part types, or sensor types, can be used to group data into the respective domains. Without loss of generality, we illustrated the three-step approach using NOTEARS-MLP. Crucially, by conducting independence tests between the residuals of \acp{mlp} trained on the pooled data and the domain indices, \ac{cicme} is able to identify variables with invariant causal mechanisms, namely common causal mechanisms that are shared by all domains. We also proposed two optimization schemes to enforce the causal structure of the stable variables to be consistently recovered while estimating the remaining causal mechanisms for each individual domain. The advantages of \ac{cicme} are demonstrated using simulation data under evaluation scenarios specifically designed to match real-world situations in manufacturing. The results presented in the paper suggest that \ac{cicme} is able to reliably find stable variables with domain-invariant causal mechanisms when provided with sufficient samples. Furthermore, it combines the advantages of both baseline methods---applying causal discovery on pooled data and on data from individual domains repeatedly, and it even outperforms both under certain scenarios. Moreover, the additional computational overhead required by \mbox{\ac{cicme}-f} is negligibly small.

The current work clearly has some limitations. The most important one lies in the prerequisite that all datasets must contain the same set of variables. Additional variables that do not appear in all domains have to be dropped. Furthermore, we observed that the HSIC test tends to falsely identify $X_3$ as unstable in \ref{exp:4}. We leave it to future work to incorporate a finetuning step using the least squares error before applying the HSIC test on the model. Another intrinsic limitation of statistical tests is that the accuracy is largely dependent on the number of available samples. When provided with too few samples, the HSIC test often fails to identify unstable variables, i.e., variables whose local causal mechanism changes across domains. Incorrectly learned common causal mechanisms can adversely impact the performance of our method. A possible direction for future research is thus to develop a method that can more reliably identify stable variables, even when the sample size is small. Ideally, such methods should also indicate the specific domains where changes occur, providing deeper insights into the behavior of the system. Moreover, it would be interesting to evaluate the behavior of CICME under different numbers of domains. Lastly, we adopted the assumption of causal sufficiency, as is common in causality research. This assumption implies that all common causes of the observed variables are also observed. However, this requirement is often violated in practice. Future research should therefore evaluate the robustness of \ac{cicme} under the violation of causal sufficiency.

\bibliographystyle{IEEEtran}
\bibliography{ref}

\begin{thebibliography}{10}
\providecommand{\url}[1]{#1}
\csname url@samestyle\endcsname
\providecommand{\newblock}{\relax}
\providecommand{\bibinfo}[2]{#2}
\providecommand{\BIBentrySTDinterwordspacing}{\spaceskip=0pt\relax}
\providecommand{\BIBentryALTinterwordstretchfactor}{4}
\providecommand{\BIBentryALTinterwordspacing}{\spaceskip=\fontdimen2\font plus
\BIBentryALTinterwordstretchfactor\fontdimen3\font minus \fontdimen4\font\relax}
\providecommand{\BIBforeignlanguage}[2]{{%
\expandafter\ifx\csname l@#1\endcsname\relax
\typeout{** WARNING: IEEEtran.bst: No hyphenation pattern has been}%
\typeout{** loaded for the language `#1'. Using the pattern for}%
\typeout{** the default language instead.}%
\else
\language=\csname l@#1\endcsname
\fi
#2}}
\providecommand{\BIBdecl}{\relax}
\BIBdecl

\bibitem{scholkopf2021toward}
B.~Sch{\"o}lkopf, F.~Locatello, S.~Bauer, N.~R. Ke, N.~Kalchbrenner, A.~Goyal, and Y.~Bengio, ``Toward causal representation learning,'' \emph{Proceedings of the IEEE}, vol. 109, no.~5, pp. 612--634, 2021.

\bibitem{mooij2016distinguishing}
J.~M. Mooij, J.~Peters, D.~Janzing, J.~Zscheischler, and B.~Sch{\"o}lkopf, ``Distinguishing cause from effect using observational data: methods and benchmarks,'' \emph{Journal of Machine Learning Research}, vol.~17, no.~32, pp. 1--102, 2016.

\bibitem{yu2024causal}
J.~Yu, T.~Pychynski, K.~S. Barsim, and M.~F. Huber, ``Causal knowledge in data fusion: Systematic evaluation on quality prediction and root cause analysis,'' in \emph{2024 27th International Conference on Information Fusion (FUSION)}.\hskip 1em plus 0.5em minus 0.4em\relax IEEE, 2024, pp. 1--8.

\bibitem{scholkopf2012causal}
B.~Sch\"{o}lkopf, D.~Janzing, J.~Peters, E.~Sgouritsa, K.~Zhang, and J.~Mooij, ``On causal and anticausal learning,'' in \emph{International Conference on Machine Learning}, 2012, p. 459–466.

\bibitem{magliacane2018domain}
S.~Magliacane, T.~Van~Ommen, T.~Claassen, S.~Bongers, P.~Versteeg, and J.~M. Mooij, ``Domain adaptation by using causal inference to predict invariant conditional distributions,'' \emph{Advances in neural information processing systems}, vol.~31, 2018.

\bibitem{arjovsky_invariant_2019}
M.~Arjovsky, L.~Bottou, I.~Gulrajani, and D.~Lopez-Paz, ``Invariant risk minimization,'' \emph{arXiv preprint arXiv:1907.02893}, 2019.

\bibitem{zheng_dags_2018}
X.~Zheng, B.~Aragam, P.~K. Ravikumar, and E.~P. Xing, ``Dags with no tears: {Continuous} optimization for structure learning,'' \emph{Advances in neural information processing systems}, vol.~31, 2018.

\bibitem{zheng_learning_2020}
X.~Zheng, C.~Dan, B.~Aragam, P.~K. Ravikumar, and E.~P. Xing, ``Learning {Sparse} {Nonparametric} {DAGs},'' in \emph{Proceedings of the {Twenty} {Third} {International} {Conference} on {Artificial} {Intelligence} and {Statistics}}, ser. Proceedings of {Machine} {Learning} {Research}, S.~Chiappa and R.~Calandra, Eds., vol. 108.\hskip 1em plus 0.5em minus 0.4em\relax PMLR, Aug. 2020, pp. 3414--3425.

\bibitem{rojas2018invariant}
M.~Rojas-Carulla, B.~Sch{\"o}lkopf, R.~Turner, and J.~Peters, ``Invariant models for causal transfer learning,'' \emph{Journal of Machine Learning Research}, vol.~19, no.~36, pp. 1--34, 2018.

\bibitem{pearl2009causality}
J.~Pearl, \emph{Causality: Models, Reasoning and Inference}, 2nd~ed.\hskip 1em plus 0.5em minus 0.4em\relax USA: Cambridge University Press, 2009.

\bibitem{peters_elements_2017}
J.~Peters, D.~Janzing, and B.~Schölkopf, \emph{Elements of causal inference: foundations and learning algorithms}.\hskip 1em plus 0.5em minus 0.4em\relax The MIT Press, 2017.

\bibitem{shimizu_linear_2006}
S.~Shimizu, P.~O. Hoyer, A.~Hyvärinen, A.~Kerminen, and M.~Jordan, ``A linear non-{Gaussian} acyclic model for causal discovery.'' \emph{Journal of Machine Learning Research}, vol.~7, no.~10, 2006.

\bibitem{hoyer_nonlinear_2008}
P.~Hoyer, D.~Janzing, J.~M. Mooij, J.~Peters, and B.~Schölkopf, ``Nonlinear causal discovery with additive noise models,'' \emph{Advances in neural information processing systems}, vol.~21, 2008.

\bibitem{peters_identifiability_2014}
J.~Peters and P.~Bühlmann, ``\BIBforeignlanguage{en}{Identifiability of {Gaussian} structural equation models with equal error variances},'' \emph{\BIBforeignlanguage{en}{Biometrika}}, vol. 101, no.~1, pp. 219--228, Mar. 2014.

\bibitem{spirtes_causation_2000}
P.~Spirtes, C.~N. Glymour, and R.~Scheines, \emph{Causation, prediction, and search}, 2nd~ed., ser. Adaptive computation and machine learning.\hskip 1em plus 0.5em minus 0.4em\relax Cambridge, Mass: MIT Press, 2000.

\bibitem{spirtes_fci_1999}
P.~Spirtes, C.~Meek, and T.~S. Richardson, ``An algorithm for causal inference in the presence of latent variables and selection bias,'' in \emph{Computation, Causation and Discovery}, C.~Glymour and G.~F. Cooper, Eds.\hskip 1em plus 0.5em minus 0.4em\relax MIT Press, 1999, ch.~6, pp. 211--252.

\bibitem{chickering_optimal_2002}
D.~M. Chickering, ``Optimal structure identification with greedy search,'' \emph{Journal of machine learning research}, vol.~3, no. Nov, pp. 507--554, 2002.

\bibitem{Lachapelle2020Gradient-Based}
S.~Lachapelle, P.~Brouillard, T.~Deleu, and S.~Lacoste-Julien, ``Gradient-based neural dag learning,'' in \emph{International Conference on Learning Representations}, 2020.

\bibitem{Zhu2020Causal}
S.~Zhu, I.~Ng, and Z.~Chen, ``Causal discovery with reinforcement learning,'' in \emph{International Conference on Learning Representations}, 2020.

\bibitem{ng_role_2020}
I.~Ng, A.~Ghassami, and K.~Zhang, ``On the role of sparsity and dag constraints for learning linear dags,'' \emph{Advances in Neural Information Processing Systems}, vol.~33, pp. 17\,943--17\,954, 2020.

\bibitem{he_daring_2021}
Y.~He, P.~Cui, Z.~Shen, R.~Xu, F.~Liu, and Y.~Jiang, ``\BIBforeignlanguage{en}{{DARING}: {Differentiable} {Causal} {Discovery} with {Residual} {Independence}},'' in \emph{\BIBforeignlanguage{en}{Proceedings of the 27th {ACM} {SIGKDD} {Conference} on {Knowledge} {Discovery} \& {Data} {Mining}}}.\hskip 1em plus 0.5em minus 0.4em\relax Virtual Event Singapore: ACM, Aug. 2021, pp. 596--605.

\bibitem{bello_dagma_2022}
K.~Bello, B.~Aragam, and P.~Ravikumar, ``Dagma: {Learning} dags via m-matrices and a log-determinant acyclicity characterization,'' \emph{Advances in Neural Information Processing Systems}, vol.~35, pp. 8226--8239, 2022.

\bibitem{deng2025markov}
C.~Deng, K.~Bello, P.~Ravikumar, and B.~Aragam, ``Markov equivalence and consistency in differentiable structure learning,'' \emph{Advances in Neural Information Processing Systems}, vol.~37, pp. 91\,756--91\,797, 2025.

\bibitem{Bertsekas01031997}
D.~P. Bertsekas, ``Nonlinear programming,'' \emph{Journal of the Operational Research Society}, vol.~48, no.~3, pp. 334--334, 1997.

\bibitem{byrd1995limited}
R.~H. Byrd, P.~Lu, J.~Nocedal, and C.~Zhu, ``A limited memory algorithm for bound constrained optimization,'' \emph{SIAM Journal on scientific computing}, vol.~16, no.~5, pp. 1190--1208, 1995.

\bibitem{gretton2007kernel}
A.~Gretton, K.~Fukumizu, C.~Teo, L.~Song, B.~Sch{\"o}lkopf, and A.~Smola, ``A kernel statistical test of independence,'' \emph{Advances in neural information processing systems}, vol.~20, 2007.

\end{thebibliography}

\end{document}